\documentclass[letterpaper, 10 pt, conference]{ieeeconf}

\IEEEoverridecommandlockouts
\usepackage[T1]{fontenc}
\usepackage{tgtermes}
\usepackage{siunitx}
\usepackage{cite}
\usepackage{url}
\usepackage[hidelinks]{hyperref}
\usepackage{graphicx}
\usepackage{amsmath}
\usepackage{amssymb}
\usepackage{balance}
\usepackage{tabularx}
\usepackage{multirow}
\usepackage{algorithm}
\usepackage{algpseudocode}
\usepackage{eso-pic}

\AddToShipoutPictureBG*{%
  \AtPageUpperLeft{%
    \raisebox{-1.3cm}{%
      \makebox[\paperwidth]{\footnotesize\sffamily
        \textbf{Accepted final version.} IEEE International Symposium on Safety, Security, and Rescue Robotics (SSRR), Incheon, South Korea, November 2026}}}}

\title{\LARGE \bf
I See You: Initial Multi-Robot LiDAR Registration with Mutual-Observation Checks
}

\author{
  Jonathan Lichtenfeld,
  Oskar von Stryk
  \thanks{\raggedright
  The authors are with the SIM Group, Technical University of Darmstadt, Germany. \texttt{\{lichtenfeld,stryk\}@sim.tu-darmstadt.de}}
  \thanks{Research presented in this paper has been supported in parts by the
LOEWE initiative (Hesse, Germany) within the emergenCITY center [LOEWE/1/12/519/03/05.001(0016)/72] 
and by the Federal Ministry of Research, Technology and Space (BMFTR) within the projects DRZ (grant no. 13N16475) and KIARA PRO (grant no. 13N17765),
and within the Hessian Drone Competence Centre Kassel Airport (HDKZ), 
funded by the European Regional Development Fund (ERDF), 
the Hessian Ministry for Digitalization and Innovation (Distr@l), 
and the Technical University of Darmstadt.
}%
}

\begin{document}
\bstctlcite{IEEEexample:BSTcontrol}
\maketitle

\thispagestyle{empty}
\pagestyle{empty}

\begin{abstract}
Compared to single-robot deployments, multi-robot teams can significantly improve search-and-rescue (SAR) mission success by covering large areas in parallel.
To collaborate effectively, these robots must be registered in a common reference frame. 
Standard registration methods typically rely on external infrastructure or significant map overlap, but real deployments require infrastructure-free, stationary initial alignment for immediate operation.
While global LiDAR point cloud registration addresses this, existing single-scan methods often fail silently in aliased or degenerate environments. 
Approaches scaling beyond two agents also remain rarely evaluated in real-world scenarios.
We propose a centralized, deployment-ready framework for pre-mission multi-robot registration.
Our system performs pairwise coarse-to-fine alignment and actively rejects impossible pose hypotheses via deterministic plausibility checks.
We validate the framework on a real-world custom dataset of four synchronized robots in various environments with ground truth.
The checks reject more than a third of the incorrect pairwise registrations while falsely rejecting less than 1\% of the correct ones, and reduce the number of team registrations with a wrong robot pose by 38\%.
Dataset and code are publicly available.
\end{abstract}

\section{INTRODUCTION}
\begin{figure}[t]
    \centering
    \includegraphics[width=\linewidth]{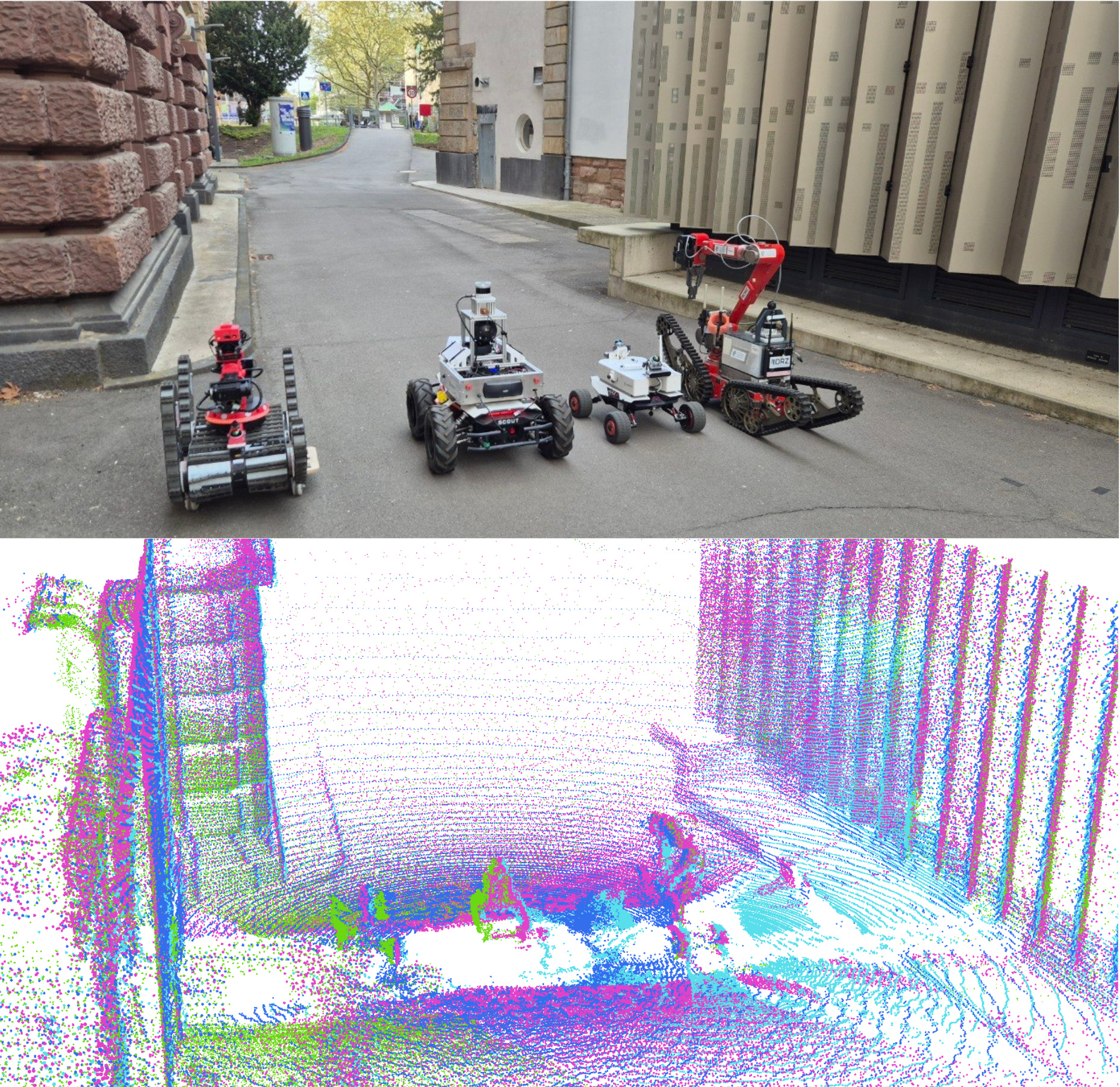}
    \caption{
        A team of four heterogeneous UGVs co-located at a common staging area (top) and their LiDAR scans aligned into a common reference frame by our method (bottom), colored per robot.}
    \label{fig:teaser}
\end{figure}

Unmanned Ground Vehicles (UGVs) accelerate search and rescue (SAR) operations by covering large disaster sites in parallel~\cite{Queralta2020,Yu2020}. 
To coordinate safely, these robots rely on multi-robot simultaneous localization and mapping (MR-SLAM). 
While MR-SLAM is a rapidly advancing field, unconditionally deploying these systems in unstructured, ad-hoc disaster zones remains a severe practical challenge.
A fundamental prerequisite for such missions is \emph{initial inter-robot registration}:
before any collaborative mapping can begin, every robot must establish its pose within a shared reference frame (Fig.~\ref{fig:teaser}).

Some MR-SLAM frameworks require a coarse initial relative pose~\cite{Yu2020}. Others attempt to build the shared frame as the robots cross paths during exploration~\cite{Ruf2025Multi,Lajoie2020door,Lajoie2024swarm}.
Yet, immediate alignment is an important requirement for field deployments.
Operators need a unified view to safely coordinate the initial rollout, and downstream SLAM algorithms benefit from a solid starting estimate to prevent early divergence.
Standard workarounds bridge this gap using manual calibration, external infrastructure, or GNSS. 
In ad-hoc, unstructured environments, these dependencies introduce critical single points of failure.

We address the common deployment scenario where robots launch from a co-located staging area. 
If their sensor fields of view (FOV) overlap, we can establish the initial reference frame immediately. 
We scope our approach strictly to gravity-aligned LiDAR data, as LiDAR provides accurate 3D geometry and serves as the primary exteroceptive sensor for modern SAR UGVs.

However, raw point cloud registration remains highly vulnerable to real-world deployment conditions.
Sparse FOV overlap, dense vegetation, or structurally repetitive architecture can easily confuse global solvers.
In these scenarios, algorithms may output completely incorrect poses with high mathematical confidence.
Because standard optimization cannot independently distinguish a physically feasible alignment from an impossible one, these results should not be directly trusted.
A deployable system requires strict plausibility checks to actively reject physically impossible alignments before the mission begins.

We present a centralized, infrastructure-free framework to achieve robust pre-mission registration between multiple UGVs using LiDAR. Our primary contributions are:
\begin{itemize}
    \item A registration framework combining global point cloud alignment with deterministic physical plausibility checks to actively reject geometrically impossible hypotheses.
    \item Real-world evaluation across diverse indoor and outdoor scenarios with a heterogeneous UGV team, demonstrating the prevention of critical pre-mission misalignments.
    \item An open-source framework implementation accompanied by a synchronized multi-robot LiDAR dataset with ground truth: \url{https://tu-darmstadt-ros-pkg.github.io/i-see-you-registration}.
\end{itemize}
\section{RELATED WORK}
\begin{figure*}[t]
    \centering
    \includegraphics[width=.85\textwidth]{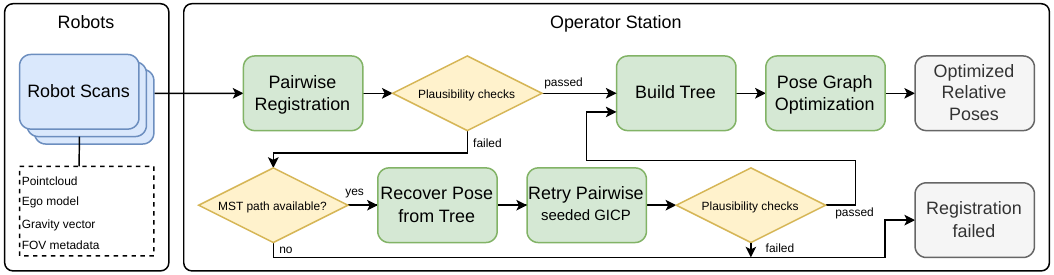}
    \caption{Overview of the proposed registration pipeline. Pairwise registrations are first filtered by physical plausibility checks. Accepted direct edges are used as candidates for a maximum-weight spanning tree, weighted by the global-registration inlier count. Rejected edges can be retried when an initial transform can be composed along the tree; if the retry passes the same checks, it is added as a verified constraint for the final pose graph optimization, but is not necessarily part of the tree itself.}
    \label{fig:framework}
\end{figure*}

\subsection{Cold-Start Multi-Robot Registration}
Establishing a shared reference frame among a team of robots is a prerequisite for any form of collaborative mapping or coordinated operation.
In the context of Search and Rescue (SAR), robots often launch simultaneously from a confined staging area.
This creates a stationary cold-start problem: the platforms must establish a unified coordinate system immediately upon deployment, before the mission begins.

\paragraph*{Pre-Placed Infrastructure and Prior Knowledge}
Methods relying on external aids guarantee high accuracy but share a fundamental drawback: they depend on setups or data that are rarely available in ad-hoc disaster scenarios.
Visual fiducial markers~\cite{tranzattoTeamCERBERUSWins2022}, for instance, provide precise pose estimates but fail under poor lighting or occlusion, and require exact camera-to-LiDAR extrinsic calibration~\cite{Xie2018}.
LiDAR-reflective calibration targets~\cite{Ruf2025Multi} resolve the illumination issue but still demand deliberate manual placement at the deployment site.
Motion capture systems~\cite{Chaney2023M3ED} yield ground truth accuracy in laboratories but are entirely impractical for field deployment.
Similarly, localization against pre-existing maps~\cite{Moura2021BIM} acts as a form of virtual infrastructure, assuming an accurate, up-to-date map exists before deployment.

\paragraph*{Infrastructure-Free Approaches}
At the most basic level, hand measurement or coarse manual placement is fast, but introduces substantial pose error with no quality guarantee.
More principled approaches cast relative pose estimation as an extrinsic calibration problem, which recovers the transform either by matching sensor trajectories over time (motion-based)~\cite{Lee2025} or by directly registering overlapping observations (geometry-based)~\cite{Harltey2006Multiple,Lindenberger2021Pixel}.
Motion-based calibration, however, requires coordinated movement across platforms, which is often impossible in a confined SAR staging area.
Geometry-based registration instead aligns simultaneously captured observations.
Visual feature registration~\cite{Lindenberger2021Pixel} matches image features across views, but is illumination-dependent and needs precise camera-to-LiDAR calibration for subsequent LiDAR-based mapping.
Geometric LiDAR registration~\cite{Kulmer2024} avoids both, but targets sensors co-mounted on a single vehicle, which cannot observe one another, as it is the case in multi-robot scenarios.
Coarse registration can additionally be aided by modalities such as RTK-GNSS~\cite{Feng2024S3E} and UWB ranging~\cite{Shule2020UWB,shalihanMRFLOURMultirobotRelative2025}; however, GNSS is typically unavailable in GPS-denied environments, and UWB requires specialized hardware that is not standard equipment on many platforms.
This work relies on multi-LiDAR registration, a geometry-based approach that provides an infrastructure-free solution leveraging standard hardware already present on most SAR platforms, while offering superior precision and illumination robustness compared to visual alternatives.

Static alignment differs from MR-SLAM~\cite{huangDiSCoSLAMDistributedScan2022,Lajoie2024swarm,Lajoie2020door}, which establishes shared coordinate frames through ongoing trajectory-based alignment and loop closure detection. 
MR-SLAM cannot guarantee a unified frame until closure is found, and the approach remains complex and difficult to deploy reliably in practice. 
Many systems benefit from accurate initial relative poses between robots, which our pre-mission registration provides.

\subsection{Multi-View Point Cloud Registration}
The registration of two point clouds is a well-studied problem, and solutions range from classical local methods such as the renowned Iterative Closest Point (ICP)~\cite{Besl1992Method} to global approaches~\cite{limKISSMatcherFastRobust2024}.

When simultaneously registering more than two point clouds, as in multi-robot scenarios, the problem scales to \textit{multi-view registration}.
A standard approach tackles this by independently computing all relative pairwise alignments~\cite{Theiler2015,SungjoonChoi2015} and organizing them into a global pose graph~\cite{SungjoonChoi2015,Wang2023} or spanning tree~\cite{Huber2003}.
As described in~\cite{Yang2024}, different strategies for spanning tree construction and edge weighting have been proposed.
Alternatively, global registration frameworks
jointly estimate all absolute poses from the set of relative transforms using similarity averaging~\cite{Cui2015}.
A method similar to our approach is~\cite{Theiler2015}, which selects the best candidates from pairwise registrations and then optimizes them in a global pose graph.
Incorrect edges can remain after that selection.
They are commonly handled by comparing edges with each other.
DOOR-SLAM rejects spurious inter-robot loop closures by pairwise consistency maximization: two loop closures are kept together only when the cycle they form with the robots' trajectories is consistent, and the largest such set is retained~\cite{mangelsonPairwiseConsistentMeasurement2018,Lajoie2020door}.
Robust pose-graph optimization instead down-weights inconsistent edges while the graph is optimized~\cite{Wang2023,yangGraduatedNonConvexityRobust2020}.
Both need several edges between the same robots, which a single snapshot of a small team often lacks.
The plausibility checks in this work therefore accept or reject each registration using only the scans of the two robots involved, before the graph is optimized.

\subsection{Registration Uncertainty and Mutual Observation}\label{sec:reg_uncertainty}
Point cloud registration is inherently uncertain and can fail due to small overlap, sensor noise, or geometric degeneracy~\cite{Gaus2025,Tuna2025}.
Registration quality is commonly summarized by the inlier count of the global registration~\cite{limKISSMatcherFastRobust2024} or by the residual of the fine alignment.
Neither a high inlier count nor a low residual guarantees a correct pose: in a symmetric environment, a registration may converge to a local minimum that is far from the true solution and still keep many inliers and a small residual~\cite{Zhang2016}.
A threshold on either score leaves these failures uncaught.

In multi-robot scenarios, mutual observation between the platforms can provide a physical constraint to validate registration hypotheses.
While relative pose consistency via visual marker detection~\cite{Chng2019} and LiDAR-based relative detection~\cite{shalihanMRFLOURMultirobotRelative2025} have been investigated, utilizing raw LiDAR line-of-sight as a deterministic rejection filter remains largely unexplored.

\section{METHOD}
We propose a robust framework for multi-robot registration that aims to detect wrong hypotheses by leveraging physical constraints.

\subsection{Assumptions and Scope}

Our framework operates under the following assumptions, which reflect the coordinated launch scenario typical of search-and-rescue missions:

\begin{itemize}
    \item \emph{Co-located initialization:} All robots start in a common area with sufficient FOV overlap; all are equipped with LiDAR and IMU and undergo gravity alignment during a static initialization phase before mission start.
    \item \emph{Team scale:} The framework is designed for small coordinated teams typical of SAR missions; evaluation and testing are conducted with four robots.
    \item \emph{Network connectivity:} Robots communicate with a central operator station via reliable wireless links; bandwidth is not a critical bottleneck within the launch area.
    \item \emph{Relaxed real-time constraints:} Strict real-time execution (e.g., $<100$\,ms) is not required; computing the alignment over several seconds is acceptable since it occurs before active robot navigation begins.
\end{itemize}

Consequently, scenarios involving late map merging and discovery (robots starting at remote locations), registration under small FOV overlap, as well as robotic swarms with dozens of agents are out of scope for this work.

\subsection{Framework}
Our framework applies pairwise registration followed by physical plausibility checks, and a global optimization of all accepted poses. Fig.~\ref{fig:framework} provides an overview of this pipeline.

\subsubsection{Optional Ego-Model Augmentation}
Since LiDAR sensors are positioned to maximize the environmental field of view, they rarely capture the robot's own chassis. 
As an optional ablation, we therefore test whether adding this missing geometry can rescue failed registrations.
We sample a point cloud from the target robot's Unified Robot Description Format (URDF) model, transform it into the local LiDAR frame, and append it to the target scan.
The first pairwise attempt remains LiDAR-only; only rejected edges are retried once with this target-side ego augmentation.
Accepted retry edges must pass the same plausibility checks, and ego-augmented clouds are not kept for the final graph optimization.

\subsubsection{Pairwise Registration}
We perform pairwise registration between all $N$ robots,
resulting in $N(N-1)/2$ runs, using a two-stage approach: first, we apply a global registration method to generate an initial alignment, followed by a local refinement step to improve accuracy.
For the global registration, we use KISS-Matcher~\cite{limKISSMatcherFastRobust2024}; its inlier count is later used as the spanning-tree edge weight.
Since the scans are gravity-aligned, we restrict the optimization to the yaw angle to speed up convergence and improve robustness.
If this yields a sufficient number of inlier features, we refine the alignment using Generalized-ICP (GICP)~\cite{Koide2024}.

\subsubsection{Plausibility Checks}
Pairwise registration may fail in ambiguous scenes or with small overlap, even though the result appears geometrically plausible.
To detect such cases, we propose a set of plausibility checks based on physical constraints, which are applied to each pairwise registration result.

\noindent\textbf{Overlapping Bounding Boxes.} If the chassis bounding boxes of the two robots overlap after registration, the result is likely incorrect, since the robots cannot occupy the same physical space.

\noindent\textbf{Mutual Observation.} If the robots have sufficient FOV overlap, they should observe each other in their LiDAR scans.
A pose is rejected when the predicted chassis box is empty and a LiDAR has sampled the space beyond that box.
This check takes into account the sensors' FOV, self-occlusions, and known bounding box dimensions of the observed robot.
Algorithm~\ref{alg:directional_observation} describes the directional verification from one observer to a target.
To verify a pairwise registration hypothesis $A \to B$, we run this check bidirectionally (evaluating A's observation of B, and B's observation of A) and reject the hypothesis if either direction is rejected.

The aggregation logic prioritizes a clear failure over an inconclusive LiDAR.
For each LiDAR in a direction: if the target is outside the field of view or occluded, that LiDAR is inconclusive;
if at least $\tau$ occupied voxels of the observer cloud lie in the target box, the direction passes;
if the box is empty and the cone beyond it contains at least $\tau_{cone}$ returns, the direction is rejected.
The cone is a beam of half-angle $\alpha$ about the ray from the LiDAR through the box center.
Returns in that beam at or beyond the box are background.
Enough of them mean the LiDAR imaged this direction, so an empty box is evidence that the target is absent.
Too few mean the direction was undersampled, and the empty box is ignored.
A single LiDAR whose cone behind an empty box contains at least $\tau_{cone}$ returns is enough to reject the hypothesis, even if the other LiDARs are occluded or outside the field of view.
If every LiDAR is inconclusive or the empty box was not sampled densely enough, the direction is accepted.
We use $\tau = 3$ occupied voxels of 5\,cm, $\alpha = 2^\circ$, and $\tau_{cone} = 5$, set empirically.

Note that neither of these checks can verify a registration hypothesis; a passed check does not necessarily mean the registration is correct, but a failed check indicates it is likely incorrect.

\begin{algorithm}
\caption{Directional Observation Check}
\label{alg:directional_observation}
\begin{algorithmic}[1]
    \Require Observer cloud $\mathcal{P}_{obs}$, Sensors $L_{obs}$, Target box $BB_{tgt}$ (in obs. frame)
    \Require Support threshold $\tau$, cone half-angle $\alpha$, cone return threshold $\tau_{cone}$
    \State $\mathcal{V} \gets \text{Voxelize}(\mathcal{P}_{obs})$
    
    \For{each sensor $l \in L_{obs}$}
        \State $r \gets \text{Ray}(l.\text{origin}, BB_{tgt}.\text{center})$
        
        \If{$r \notin l.\text{FOV}$ \textbf{or} $\text{IsOccluded}(r, \mathcal{V})$}
            \State \textbf{continue} \Comment{Target not visible; cannot falsify}
        \EndIf
        
        \If{$\text{CountOccupiedVoxels}(\mathcal{V}, BB_{tgt}) \ge \tau$}
            \State \Return \texttt{Accept} \Comment{Target observed}
        \EndIf
        
        \State $b \gets \text{CountReturnsInConeBeyond}(\mathcal{P}_{obs}, r, \alpha, \|r\|)$
        
        \If{$b \ge \tau_{cone}$}
            \State \Return \texttt{Reject} \Comment{Space was sampled and is empty}
        \EndIf
    \EndFor
    
    \State \Return \texttt{Accept} \Comment{Observed, occluded/out of FOV, or undersampled}
\end{algorithmic}
\end{algorithm}

\subsubsection{Maximum-Weight Spanning Tree and Pose Graph Optimization}
\begin{figure}[t]
    \centering
    \includegraphics[width=\columnwidth]{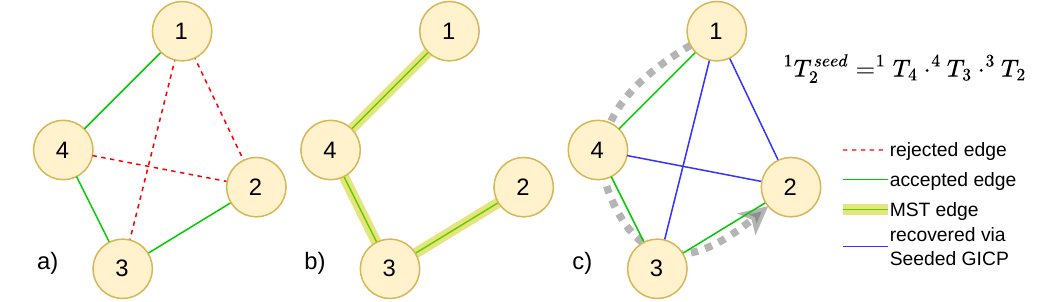}
    \caption{Maximum-weight spanning tree:
     Initially, only a subset of pairwise registrations succeeds, while others fail (a).
     The tree is extracted from the successful edges (b). It allows deriving initial guesses for the failed edges (c), which are then re-computed and checked.
     }
    \label{fig:mst}
\end{figure}

The accepted pairwise poses define a graph that may contain cycles and redundant edges.
We extract a maximum-weight spanning tree, or forest if the verified graph is disconnected, using the global-registration inlier count as edge weight (Fig.~\ref{fig:mst}).
The tree enables recovery of rejected loop closures by seeding failed registrations with poses inferred along its paths.
If seeded refinement converges and passes plausibility checks, the edge is reinstated as a verified pose-graph constraint.

Finally, all verified edges are assembled into a pose graph and optimized using GTSAM~\cite{Dellaert2022} to produce globally consistent alignment.
Each edge is weighted by an isotropic noise model whose standard deviation grows with the GICP residual of the registration. 
A Huber kernel is used to reduce the influence of large residuals.

For a team of only two robots, the spanning-tree and pose graph optimization steps are omitted, since they would not provide additional information.
In this case, the pairwise registration result is directly used as the final alignment, or rejected if the plausibility checks fail.
\section{DATASET}
Existing multi-UGV LiDAR datasets remain limited in several respects.
Many span only a narrow range of environments~\cite{SaeediCOBRA} and use homogeneous fleets of identical platforms~\cite{Tian2021KimeraMulti,Feng2024S3E,Leung2011} or sensors~\cite{Zhou2024Coped,Chaney2023M3ED}, often without accurate ground truth~\cite{Tian2021KimeraMulti,Zhou2024Coped}.
Others deliberately minimize mutual visibility to study map exchange upon encounter, or are assembled from independent single-robot sequences recorded in the same environment rather than synchronous multi-robot captures~\cite{Albin2025CUMulti}.
In general, SLAM datasets feature moving robots, which introduces motion distortion into the LiDAR scans, whereas static scenarios are rare.
As a result, realistic inter-robot observations and occlusions remain underrepresented.

\begin{figure}
\includegraphics[width=\columnwidth]{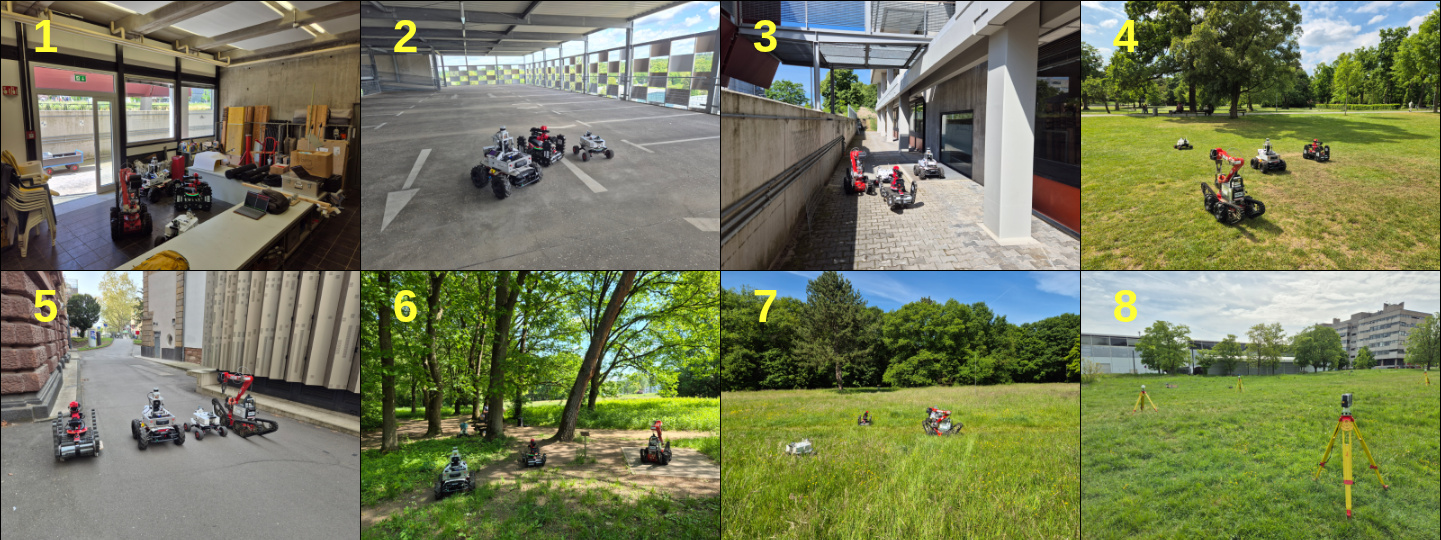}
\caption{Overview of the captured scenarios. 1: Lab, 2: Parking, 3: Walkway, 4: Campus, 5: Forest, 6: Meadow, 7: Park, 8: Capture of TLS reference scans.}
\label{fig:dataset_overview}
\end{figure}

To this end, we collected a comprehensive dataset of multi-robot LiDAR scans in various indoor and outdoor settings (see Table \ref{tab:dataset_summary}, Fig.~\ref{fig:teaser},~\ref{fig:dataset_overview}).
The platform selection mirrors the diversity of real-world multi-robot teams, including both small and large platforms with different mobility types (wheeled platforms for quick exploration, and tracked platforms equipped with manipulators for uneven terrain).

\begin{figure}
\includegraphics[width=\columnwidth]{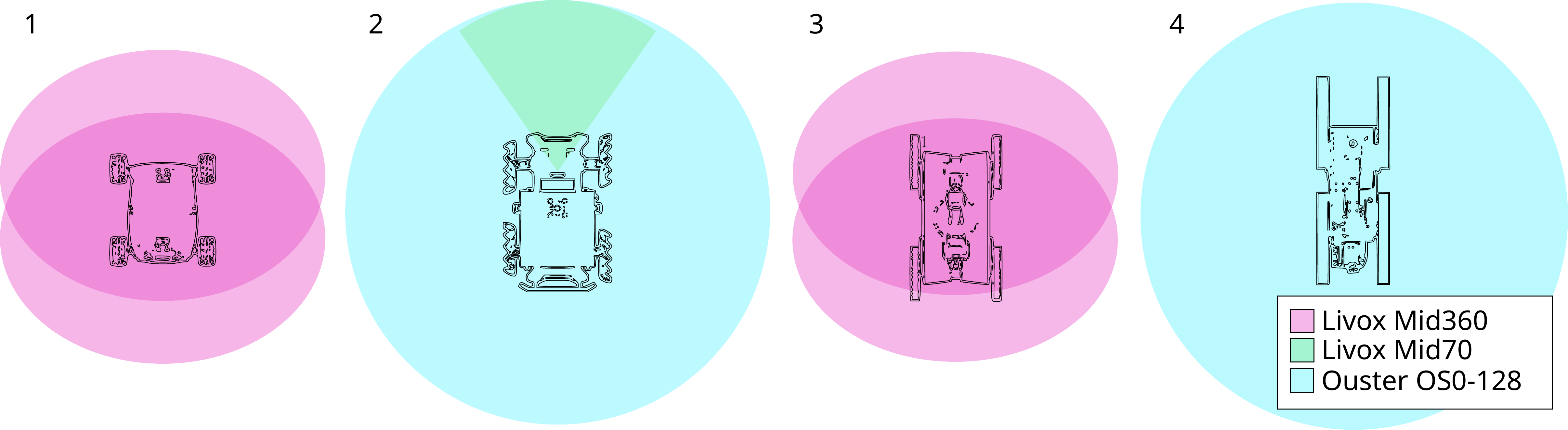}
\caption{Field of view of the different LiDAR sensors mounted on the platforms, as listed in Table~\ref{tab:platforms}.
The Livox Mid360 units are tilted to also cover the ground in front of their respective side of the robot and increase their joint FOV, while the other sensors are positioned horizontally.
Self-occlusion is not shown for platforms 3 and 4; it depends on the current configuration of the tracks and manipulator.}
\label{fig:fov}
\end{figure}

Within each scene, we captured multiple static constellations of the robots (in the following called \textit{snapshots}), including varying degrees of mutual visibility, occlusion and FOV overlap.
This ranges from close and aligned formations at a launch site to more dispersed arrangements, with and without line-of-sight between the robots.
All robots were synchronized via NTP.
The platforms are equipped with heterogeneous off-the-shelf LiDAR sensors, including an Ouster OS0-128, Livox Mid-360, and Livox Mid-70, which provide diverse scanning patterns and fields of view (see Table \ref{tab:platforms}, Fig.~\ref{fig:fov}).
To account for the sparsity of the Mid-360 and Mid-70, we accumulated LiDAR scans over 1s, leading to $\sim$220k points per scan for the Mid-360 and $\sim$100k points for the Mid-70, opposed to $\sim$131k points for the OS0-128.
The sensors' built-in IMUs were sampled for 1s and the mean gravity vector was stored along with the LiDAR scan, to allow for gravity-alignment.
The Mid-70 does not have an IMU, but is mounted with the same orientation as the OS0-128, so we can use the OS0's gravity vector for both sensors on the same platform.

For ground-truth poses, we captured the scenes using a survey-grade terrestrial laser scanner (TLS).
All scans were fine-registered against the TLS map using a combination of manual initialization and GICP refinement.
We deliberately registered all scans from one platform independently, since their relative transforms might change slightly between snapshots due to the robots' mobility and non-rigid mounting of the sensors.
Since the scans are aligned directly to the TLS map, the ground truth avoids the imprecise marker-to-sensor calibration required by motion-capture or total-station setups; its remaining error stems mainly from the sensor noise and the scan-to-map registration.
The resulting ground truth poses are accurate to the order of a few centimeters, sufficient for evaluating initial registration.
For the Forest, Meadow and Park scenes, the TLS method was not feasible due to vegetation growth and the lack of distinct features; 
instead, we carefully registered the scans to each other using GICP to obtain a consistent reference frame for evaluation, albeit less accurate.

\begin{table}[ht]
\centering
\caption{Summary of the collected multi-robot LiDAR dataset.}
\label{tab:dataset_summary}
\begin{tabular}{l c c c l}
\hline
Scene & Setting &Snapshots & Platforms & GT Ref.\\
\hline
Lab & Indoor & 7 & all & TLS \\
Parking & Indoor & 9 & 1,2,3 & TLS \\
Walkway & Semi-outdoor & 13 & all & TLS \\
Campus & Outdoor & 15 & all & TLS \\
Forest & Outdoor & 6 & all & GICP \\
Meadow & Outdoor & 7 & all & GICP \\
Park & Outdoor & 12 & 1,2,3 & GICP \\
\hline
\end{tabular}
\end{table}

\begin{table}[ht]
\centering
\caption{Overview of the used platforms (cf. Fig.\ref{fig:teaser} top).}
\label{tab:platforms}
\begin{tabular}{c c c c}
\hline
\# & Platform & LiDAR & Occlusion \\
\hline
1 &Wheeled, small & $2\times$ Mid-360 & none \\
2 &Wheeled, medium & OS0-128, Mid-70 & none \\
3 &Tracked, small & $2\times$ Mid-360 & Man., Tracks \\
4 &Tracked, large & OS0-128 & Man., Tracks \\
\end{tabular}
\end{table}

\section{EVALUATION}
We evaluate our framework on the custom dataset.
Note that we do not focus on the accuracy of the pairwise registration itself, as this is not the main contribution of our work, and we rely on existing methods for this step.
Instead, we assess the overall framework's ability to produce accurate and consistent multi-robot registrations, with a particular focus on the effectiveness of the proposed plausibility checks and the final optimized registration.
Nevertheless, we report the success rate of the pairwise registration as a baseline for comparison.

\subsection{Setup}
We run the pipeline on every snapshot, enumerating all combinations of participating robots and their LiDARs.
That is, for each snapshot, we consider all possible constellations of robots, and either the first, second, or both LiDARs on each robot, depending on the platform.
Since a pairwise registration does not depend on the other robots in the constellation, each unique combination of robot pair and LiDAR selection is registered and evaluated once.
This allows us to evaluate the plausibility checks across a wide range of inter-robot configurations, differing in distance, relative orientation, FOV overlap, and mutual visibility.

All tests were run on Ubuntu 24.04.4 LTS with an Intel Core Ultra 9 285H (16 cores), and 64\,GB RAM.

\subsection{Results}
We first define the protocol and metrics, then report the effect of each component in turn.
A pairwise registration is \emph{correct} if its translation and rotation error against the ground truth stays below 20\,cm and 5$^\circ$; otherwise it is \emph{incorrect}. 
This conservative threshold is considered sufficient for initial alignment.

\noindent\textbf{Plausibility Checks.}
Following the setup above, we enumerate all robot--LiDAR constellations on every snapshot and evaluate each pairwise registration.
From all incorrect pairwise registrations, i.e., those that exceed the mentioned thresholds, we evaluate the amount of rejected or falsely accepted ones by the plausibility checks.
As shown in Table~\ref{tab:checks_breakdown_extended}, the bounding-box overlap check rejects 64 incorrect registrations, and the mutual observation check rejects another 136 of the remaining incorrect registrations.
Together, the checks reject 200 of 536 incorrect direct registrations (37.3\%), and falsely reject 7 of the 1145 correct ones (0.6\%).

The bounding-box check is an exact geometric predicate: given the estimated poses and box dimensions, overlap is computed analytically and is not affected by sensor noise or point sampling.
False rejections can still occur with robots in close proximity or imprecise bounding box dimensions. In this dataset the check rejects two LiDAR selections of one close pair.

The mutual observation is more complex: it can be affected by sensor noise, downsampling of point clouds, and other objects detected at the predicted location.
For Table~\ref{tab:los_check_per_scene}, we consider only pairs on which the check can decide: at least one robot has the other within its FOV with clear line of sight, and the pair was not already rejected by the bounding-box check.
On these pairs, it rejects a pose only when the predicted chassis box contains fewer than $\tau$ occupied voxels, and the cone beyond the box was sampled (Algorithm~\ref{alg:directional_observation}).
That condition is met when a wrong pose places the target in observed free space.
In our data this means open, roughly planar scenes with long visibility: the true-positive rate is 0.70 on the Walkway and 0.71 in Parking, with no false rejections.
In structure-rich or cluttered scenes (lab, meadow) the check rarely fires.
Overall, the mutual observation check accounts for the largest share of caught failures.

\begin{table}[ht]
\centering
\caption{Pairwise registrations through the matcher and the plausibility checks. Wrong/correct with respect to the ground truth (20\,cm, 5$^\circ$); percentages: cumulative share of the wrong pairs rejected.}
\label{tab:checks_breakdown_extended}
\begin{tabular}{l c c}
\hline
Stage & Wrong & Correct \\
\hline
All pairwise registrations & \multicolumn{2}{c}{1927} \\
Rejected by matcher (too few inliers) & \multicolumn{2}{c}{246} \\
Accepted by matcher & 536 & 1145 \\
\hline
Rejected by bounding-box check & 64 (11.9\%) & 2 \\
Rejected by mutual-observation check & 136 (37.3\%) & 5 \\
\hline
Passed both checks & 336 & 1138 \\
\hline
\end{tabular}
\end{table}

\begin{table}[ht]
\centering
\caption{Mutual-observation check per scene (clear line of sight, bounding-box check passed). TPR / FPR: share of incorrect / correct pairs rejected.}
\label{tab:los_check_per_scene}
\begin{tabular}{l c c c}
\hline
Scene & \#Incorrect & TPR $\uparrow$ & FPR $\downarrow$ \\
\hline
Lab      & 14  & 0.07 & 0.0 \\
Parking  & 42  & 0.71 & 0.0 \\
Walkway  & 122 & 0.70 & 0.0 \\
Campus   & 15  & 0.40 & 0.02 \\
Forest   & 19  & 0.16 & 0.0 \\
Meadow   & 93  & 0.03 & 0.0 \\
Park     & 26  & 0.27 & 0.0 \\
\end{tabular}
\end{table}

A closer look at the False Negatives, i.e., the incorrect registrations that passed the checks, shows that the naive check for presence of points around the predicted location of the target robot can be fooled by other objects in the scene, such as vegetation or other robots.
A translational registration error that exceeds the threshold applied for evaluation can still be small enough to keep a portion of the target robot within the FOV of the observer, leading to a false acceptance.

\noindent\textbf{Comparison with Registration Scores.}
We compare our checks with two registration scores: the KISS-Matcher inlier count of the global registration and the root-mean-square error (RMSE) of the final alignment over point pairs closer than 0.5\,m, computed post hoc on the stored registrations.
With each cutoff set to reject the same 200 wrong pairs as the checks, both scores reject more correct pairs (Table~\ref{tab:score_vs_checks}).
The checks and the RMSE also catch different failures:
the checks mostly reject poses with a correct rotation but an offset of several meters (median $0.7^\circ$, 8.0\,m), which the RMSE cannot detect since the shifted scans still overlap locally (median RMSE 0.20 vs.\ 0.18\,m for correct pairs).
The RMSE mostly flags rotation errors (median $59^\circ$).
Since the two reject largely different wrong pairs (47 of 200 in common), combining them is promising and left for future work.

\begin{table}[ht]
\centering
\caption{Score cutoffs chosen to reject as many wrong pairs as the plausibility checks, on matcher-accepted pairs.}
\label{tab:score_vs_checks}
\setlength{\tabcolsep}{3pt}
\begin{tabular}{l c c c}
\hline
Rule & Cutoff & Wrong rejected & Correct rejected \\
\hline
Plausibility checks & -- & 200 (37.3\%) & 7 (0.61\%) \\
KISS-Matcher inliers & $< 11$ & 200 (37.3\%) & 130 (11.4\%) \\
RMSE ($<0.5$\,m) & $\geq 0.249$\,m & 200 (37.3\%) & 11 (1.0\%) \\
\hline
\end{tabular}
\end{table}

\noindent\textbf{Pipeline Registration.}
For end-to-end evaluation, we run one verified MST+PGO registration per snapshot with all robots present and combined LiDAR (69 snapshots), without enumerating constellations.
Table~\ref{tab:pipeline_summary} reports the end-to-end outcome; pairwise check rates are given above.
A run succeeds only if every robot is placed and every final pose is correct; a failed run either places at least one robot at a wrong pose or leaves at least one robot unregistered.
Mean and median pose errors use TLS-referenced scenes only (Table~\ref{tab:dataset_summary}).
In 33 snapshots the matcher admitted at least one GT-incorrect direct pair; in 7 of these (21.2\%) all bad pairs were caught by the checks and the team still aligned within GT.

\begin{table*}[t]
\centering
\caption{Full-team pipeline summary (one run per snapshot, all robots, combined LiDAR).}
\label{tab:pipeline_summary}
\begin{tabular}{l l l}
\hline
\textbf{Stage / Metric} & \textbf{Result} & \textbf{Evaluated over} \\
\hline
Snapshots evaluated &
69 &
One verified MST+PGO run per snapshot \\

Full team success (GT) &
51 / 69
(\textbf{73.9\%}) &
All robots placed; all final team-pair poses within thresholds \\

Wrong final pose &
14 / 69
(20.3\%) &
$\geq$1 placed robot outside thresholds \\

Not all robots placed &
4 / 69
(5.8\%) &
No wrong pose, but $\geq$1 robot left unregistered \\

Mean final pose error &
\textbf{5.40 cm / 0.51$^\circ$ mean} &
TLS-referenced scenes only (Tab.~\ref{tab:dataset_summary}); 35 GT-successful runs \\

Median final translation error &
\textbf{4.86 cm} &
same runs \\

At-risk snapshots (GT-wrong pair KM-accepted) &
33 / 69
(47.8\%) &
Snapshots with $\geq$1 bad constraint admitted without checks \\

All bad pairs caught, team recovered &
7 / 33
(\textbf{21.2\%}) &
At-risk snapshots: checks blocked every bad edge; team within GT \\

Mean pipeline wall time &
5.60\,s &
All runs \\
\hline
\end{tabular}
\end{table*}

\noindent\textbf{Comparison with Robust Back Ends.}
Robust pose-graph optimization is the common alternative for removing wrong edges.
We therefore run the pipeline on every robot subset of every snapshot (2--4 robots, combined LiDAR), with the plausibility checks off or on, and with two different back ends: the default Huber kernel and graduated non-convexity with a truncated least-squares loss (GNC-TLS)~\cite{yangGraduatedNonConvexityRobust2020}.
With the checks off, every edge that the matcher accepts enters the graph, and rejected edges are not retried along the tree.
With the noise scaled by the GICP residual, edge errors rarely exceed the Huber threshold, so the Huber back end gives the same outcome as plain least squares in all 69 full-team runs.
We therefore report only the Huber variant.
GNC-TLS does not reduce wrong poses (Table~\ref{tab:baselines}) since there is no pose graph with two robots, and with three robots a wrong edge contradicts the other two but cannot be identified.
Pairwise consistency maximization~\cite{mangelsonPairwiseConsistentMeasurement2018,Lajoie2020door} faces the same limit: it closes cycles through the robots' trajectories, and with a single snapshot per robot only the cycles among the inter-robot edges remain.
The checks judge each edge from the two scans alone.
With the Huber back end they cut wrong poses from 127 of 598 to 79, and correct teams rise from 424 to 432 (Table~\ref{tab:baselines}).
Most of the removed failures leave a robot unregistered, which is the safer outcome before a mission.
Combined with the checks, GNC-TLS gives no further improvement.

\begin{table}[ht]
\centering
\caption{Checks and robust back ends.}
\label{tab:baselines}
\scriptsize
\setlength{\tabcolsep}{3pt}
\begin{tabular}{l r c c c c}
\hline
 & & \multicolumn{2}{c}{Checks off} & \multicolumn{2}{c}{Checks on} \\
Team & Runs & Huber & GNC-TLS & Huber (ours) & GNC-TLS \\
\hline
2 robots & 345 & 252 / 54 & 252 / 54 & 249 / 30 & 249 / 30 \\
3 robots & 207 & 143 / 56 & 135 / 64 & 150 / 37 & 146 / 41 \\
4 robots & 46  & 29 / 17  & 27 / 19  & 33 / 12 & 31 / 14 \\
\hline
All      & 598 & 424 / 127 & 414 / 137 & 432 / 79 & 426 / 85 \\
\hline
\end{tabular}\\[2pt]
{\footnotesize Correct runs / runs with a wrong final pose. The others leave a robot unregistered.}
\end{table}

\noindent\textbf{Ego-Model Augmentation.}
We evaluate the optional ego-model variant as an ablation.
Although pairwise effects varied by scene, including several structure-poor snapshots where ego geometry corrected individual pair registrations, these gains did not translate into a reliable full-team improvement.
Across all paired full-team trials, the variant changed only 11 outcomes: 6 runs were recovered and 5 were degraded.
Moreover, only 34 of 235 accepted ego-retry constraints were correct with respect to ground truth.
Hence, we exclude ego augmentation from the final pipeline; all results above use the LiDAR-only variant.
The per-scene gains in structure-poor scenes suggest that the idea is worth revisiting.

\noindent\textbf{Bandwidth \& Runtime.}
The transmitted data per robot does not exceed 2094\,KB per snapshot when applying point cloud compression with 1\,cm precision.
By applying a voxel grid filter on the robot side, this could be further reduced.
The overhead of the gravity vector and FOV information is negligible.

Across all evaluated LiDAR-only full-team runs, the complete strategy takes 5.60\,s on average.
Runtime is scene-dependent: open or vegetation-heavy scenes (meadow, forest) require longer registrations, while structured environments (parking, walkway) are faster.
It also depends on the number of robots, the number of recovered pairs, and parameter choices such as the GICP convergence criteria.
In particular, the all-pairs registration stage scales quadratically with the number of robots and therefore becomes the dominant limitation for larger teams.
For a one-time pre-mission setup, this is acceptable, but for repeated registration during a mission, the runtime could be reduced by using a faster pairwise registration method or parallelizing the pairwise registrations.

To demonstrate the field applicability of our approach, we have successfully tested a ROS 2 implementation online with a standard WiFi connection between the robots and the operator station.

\section{CONCLUSION AND OUTLOOK}

We presented a novel LiDAR-based multi-robot registration framework for coordinated search-and-rescue deployment scenarios. 
The framework combines pairwise registration with physical plausibility checks and pose graph optimization, reducing the impact of incorrect pairwise hypotheses while preserving correct ones.
Experiments on a custom dataset captured with a real multi-robot team show that the proposed checks reject more than a third of the wrong direct registrations with a false-rejection rate below 1\%, and that the final optimized poses achieve sufficient accuracy in successful runs.

Several limitations remain. 
The mutual-observation check rejects gross errors mainly in open, planar scenes. In clutter and vegetation it usually abstains, because other objects occupy the predicted volume or block the line of sight. It also cannot reject all small translational or rotational errors, and it applies only when the robots can observe each other.
As with any point-cloud registration pipeline, the pairwise stage still depends on sufficient geometric overlap and sensor coverage, and the checks on known chassis dimensions and calibrated LiDAR extrinsics.
The single-snapshot setup is intentional for confined SAR staging areas, where moving robots before deployment can be undesirable.
However, where such motion is acceptable, short local submaps could provide additional information to refine the registration over time while retaining the plausibility checks, moving the initial registration toward multi-robot SLAM.
Lastly, the presented work targets coordinated ground-robot teams rather than large swarms.
The method can be applied beyond the four-robot teams used in our evaluation, but the all-pairs registration stage becomes the runtime bottleneck.
The underlying approach is not limited to ground robots in principle, but extending it to aerial platforms would require relaxing the static-snapshot assumption and explicitly handling motion during data acquisition.
\section*{ACKNOWLEDGEMENTS}
We would like to thank Anna Sviridova, Aaron Wickert and Finn Winter for their support in data collection.

\bibliographystyle{IEEEtran}
\bibliography{bibliography}

\end{document}